\documentclass[oribibl]{llncs}

\usepackage{llncsdoc}
\usepackage{graphicx}
\usepackage[numbers]{natbib}
\usepackage{mathtools}
\usepackage[outline]{contour}
\usepackage{xcolor}
\usepackage{amssymb}
\usepackage{etoolbox}
\usepackage{multirow}

\begin{document}

\title{Tramp Ship Scheduling Problem with Berth Allocation Considerations and Time-dependent Constraints}

\titlerunning{Tramp Ship Scheduling Problem with Berth Allocation Considerations \ldots}

\author{Francisco L\'opez-Ramos \inst{1,2} \thanks{Corresponding author Tel.: +52-181-22536444}\and Armando Guarnaschelli\inst{2} \and Jos\'e-Fernando Camacho-Vallejo\inst{3} \and Laura Hervert-Escobar\inst{1} \and Rosa G. Gonz\'alez-Ram\'irez \inst{4}}

\institute{Instituto Tecnol\'ogico y de Estudios Superiores de Monterrey, Monterrey, M\'exico\\
\email{\{francisco.lopez.r, laura.hervert\}@itesm.mx}
\and
Pontificia Universidad Cat\'olica de Valpara\'iso, Valpara\'iso, Chile\\
              \email{\{francisco.lopez.r, armando.guarnaschelli\}@pucv.cl}
              \and
              Universidad Aut\'onoma de Nuevo Le\'on, San Nicol\'as de los Garza, M\'exico\\
\email{jose.camachovl@uanl.edu.mx}
\and
              Universidad de Los Andes, Santiago de Chile, Chile\\
\email{rgonzalez@uandes.cl}
}


\maketitle

\begin{abstract}
This work presents a model for the Tramp Ship Scheduling problem including berth allocation considerations, motivated by a real case of a shipping company. The aim is to determine the travel schedule for each vessel considering multiple docking and multiple time windows at the berths. This work is innovative due to the consideration of both spatial and temporal attributes during the scheduling process. The resulting model is formulated as a mixed-integer linear programming problem, and a heuristic method to deal with multiple vessel schedules is also presented. Numerical experimentation is performed to highlight the benefits of the proposed approach and the applicability of the heuristic. Conclusions and recommendations for further research are provided.


\end{abstract}

\section{Introduction}

Maritime shipping accounts for more than 80\% of foreign trade, positioning this mode of transport as the most relevant in world trade. The world fleet has experienced a continuous growth with an annual increase of 4.1 percent in 2013, reaching a total of 1.69 billion dwt (deadweight tonnages) in January 2014 \citep{UNCTAD04}. Shipping companies are constantly seeking to achieve economies of scale in order to reduce transportation costs. These costs are highly influenced by fuel cost. Because of this, the planning and scheduling of routes is critical.

The ship scheduling problem considering berth assignment has been pointed out as a promising area of research in both operational and economic terms \cite{BM15, CFNR13}. These issues have practical implications especially for the shipping companies that do not have control of berth assignment decisions at public port terminals.

In this paper, we address a tramp ship scheduling problem (TSSP) with berth allocation considerations and time-dependent constraints. The problem is motivated by a tramp shipping company that provides transport services to export palletized fruit. The aim of the decision problem is to determine the route for each vessel, the operations and travel schedule whilst satisfying berth access limitations which are explicitly modeled. Due to the access limitations to berths (availability and draft restrictions), the shipping company is allowed to assign a vessel to the same berth several times to overcome capacity and operational limitations. A mixed-integer linear programming model using an expanded berth-node network representation is proposed, as well as a heuristic decomposition method to deal with large-sized instances.

The remainder of the paper is organized as follows. Section \ref{sRW} discusses the related works on TSSP and Berth allocation problems. Section \ref{sPS} describes the problem under study. Section \ref{sNF} presents the underlying network structure. Section \ref{sNF} introduces the notation and the mathematical formulation of the optimization model. Section \ref{sSSA} provides the solution approach. Section \ref{sRP} presents the numerical results. Finally, the conclusions and recommendations for further research are provided in Section \ref{sCFR}.

\section{Related works}\label{sRW}

This work is related to two well-known problems, the Ship Scheduling Problem and the Berth Allocation Problem (BAP). The integration of these two intertwined problems has recently been pointed out as a promising area of research in both operational and economic terms \cite{BM15,CFNR13}.

In this regard, the work in \cite{LP11} proposes an integrated model for ship scheduling and berth allocation, motivated by the case of shipping lines with self-owned terminals. In a follow-up work \cite{PL14}, the authors incorporate transshipment. Both works present a case study of a feeder service company operating around the Pearl River Delta region. A discrete layout of each port terminal with time window constraints is considered. However, a vessel cannot visit the same terminal more than once during the planning horizon.

The problem addressed in this work is concerned with the transportation of palletized perishable cargo (fruits) during the fruit season. There is a wide variety of industrial applications (e.g. \cite{HFKN12, JK04}) related to the bulk size and break-bulk cargo. However, this work differs from other approaches by considering operational features such as decisions related to the number of contracts to be fulfilled, load splitting at berths instead of ships \cite{FR13}, \cite{KF10, KFL11}. It also includes features from the BAP, such as spatial, temporal, handling time, and performance \cite{BM15}. The spatial attribute describes the berth layout and water depth restrictions. Temporal attribute are constraints for providing the service at berths. The handling time attribute applies for vessel and berths. Finally, the performance attribute is associated with the objective function.

Under these considerations, this work contributes to the state-of-art on TSSP by considering specific characteristics originated by a practical case study. Therefore, filling a gap in the literature taking into account real-world features of the problem \cite{FR13}.

\section{Problem definition}\label{sPS}

Consider a set of contracts $C$ with due date, destination and cargo to be fulfilled by the shipping company. There exists a number of port terminals for loading the cargo in the vessels. Each port terminal has a number of berths with different draft capacities limiting the amount of cargo that can be loaded in the vessels. Also, berths have a variety of time windows in which vessels can be moored.

A consideration in the problem is that cargoes may not be fully loaded in a vessel if not profitable. If a cargo is not fully loaded, a compensation expense is paid to its customer.  Also, the draft of the vessel should be evaluated during the definition of the schedule. The draft of a vessel is equivalent to the maximum displacement when fully loaded. A vessel-displacement takes place whenever there is a change in the vertical distance between the waterline and the bottom of the hull when loading pallets in the vessel. In this way, it is desired to find:
\begin{itemize}
  \item A route for each vessel, composed of the ordered set of berths to be visited.
  \item A schedule for each route, considering the arrival and departure times at time windows associated with the berths contained in the route and avoiding time clashes among vessels operating in the same time window.
  \item Amount of pallets to be loaded in each vessel in each berth time window in which the vessel operates.
\end{itemize}

The problem is modeled as a mixed-integer linear programming model, with the following assumptions:

\begin{itemize}
  \item Cargoes are pre-assigned to one or more vessels.
  \item Cargoes are loaded in vessels at a constant average rate.
  \item Vessels origins are known beforehand.
  \item Vessels destinations are associated with the destination ports of their pre-assigned cargoes.
  \item Vessels have enough capacity to carry all their assigned cargoes.
  \item Vessels sail with a specific constant speed on each leg.
  \item Vessels have enough fuel in order to reach their destination.
\end{itemize}

\section{Network flow representation}\label{sNF}

Cargoes are divided into pallets that are represented as variable flows going through a directed graph $G=(N,A)$. Nodes $i \in N$ contain the vessel origins, the berth time windows in which pallets can be loaded in the vessels, and the vessel destinations according to the pre-assigned cargoes; whereas, links $(i,j) \in A$ model possible sailing legs.

\begin{figure}[h!]
\begin{center}
\centerline{{\scalebox{0.8}{\includegraphics[width=13 cm,height=7 cm]{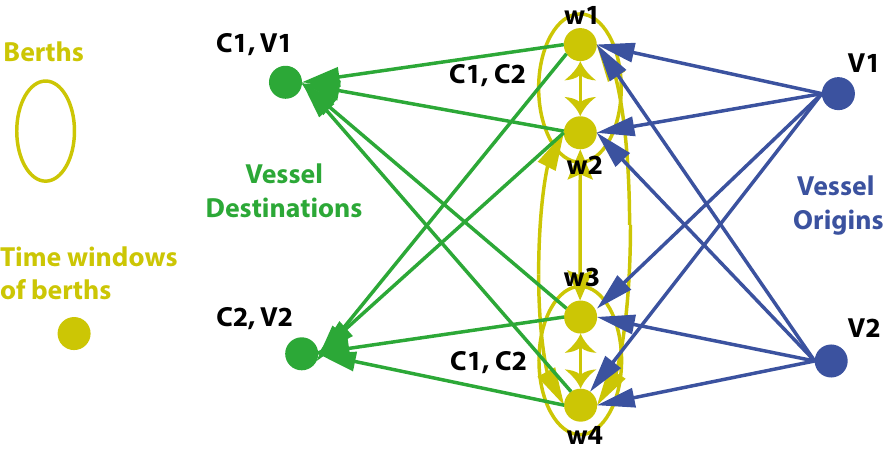}}}}
\caption{The network flow representation for a case consisting of two vessels, two contracts (one per vessel), and two berths, having each one two time windows ($w_{i}, w_{j}$).}\label{fNet}
\end{center}
\end{figure}

An illustrative example consisting of two vessels, two contracts (one per vessel) and two berths, having each one two time windows ($w_{i}, w_{j}$), is shown on Fig. \ref{fNet}. In that example, vessels start their routes at the blue nodes representing the vessel origins, and sail to some yellow nodes reproducing the available time windows of berths where pallets from cargoes C1 and C2 can be loaded in vessels V1 and V2, respectively. Vessels may dock at the same berth twice (one in each time window) if necessary. Time windows of different berths can also be used by the same vessel. This practice is frequently used by the shipping company as it allows reducing operation costs significantly. Loaded pallets from cargoes are finally carried to the green nodes, representing their destinations established in the contract.

\section{Model formulation}\label{sMF}

This section presents the mathematical formulation of the optimization model for the tramp ship scheduling problem with berth allocation issues.

\subsection*{Decision variables}

\begin{description}
  \item[$d_{ij}^{v}$] Continuous non-negative variable denoting the draft variation of vessel $v \in V$ when sailing through link $(i,j) \in \tilde{A}(v)$.
   \item[$p_{w}^{c}$] Continuous non-negative variable denoting the number of pallets loaded in vessel $v \in V$ from contract $c \in C^v$ in berth time window $w \in W^v$.
  \item[$s^{c}$] Continuous non-negative variable denoting the number of pallets from contract $c \in C$ which are not transported by any vessel.
  \item[$t_{i}^{v}$] Continuous non-negative variable denoting the arrival time of vessel $v \in V$ at node $i \in N(v)$.
  \item[$x_{ij}^{v}$] Binary variable with value 1 when vessel $v \in V$ sails through link $(i,j) \in A(v)$, value 0 otherwise.
  \item[$y_{w}^{v_{1},v_{2}}$] Binary variable with value 1 when vessel $v_{1} \in V$ docks before vessel $v_{2} \in V$ in time window $w \in W^{v_{1},v_{2}}$, value 0 otherwise.
 \end{description}

\subsection*{Optimization Model}


\begin{flalign}\label{FO}
  \max_{\boldsymbol{d}, \boldsymbol{p}, \boldsymbol{s}, \boldsymbol{t}, \boldsymbol{x}, \boldsymbol{y}} \; \; f = \!\!\!\! \sum_{{\scriptsize \begin{array}{l} v \in V\\ c \in C^v\\ w \in W^{v} \end{array}}} \!\!\!\! \psi_{w}^{c} p_{w}^{c} \; - \!\!\!\! \sum_{{\scriptsize \begin{array}{c} v \in V\\ w \in W^{v}\\i \in N^w\\ j \in N_{i}^{+}(v) \end{array}}} \!\!\!\!\!\! \theta_{w}^{v} x_{ji}^{v} \; - \!\! \sum_{{\scriptsize \begin{array}{c} v \in V\\j \in D(v)\\i \in O(v)\end{array}}} \!\!\!\! \rho^{v} \left(t_{j}^{v} - t_{i}^{v}\right)
\end{flalign}\vspace{-0.2in}
\begin{flalign}\nonumber
   \hspace{25mm} - \!\! \sum_{{\scriptsize \begin{array}{c} v \in V \\(i,j) \in A(v) \end{array}}} \!\!\!\!\!\!\!\!  \phi^{v} \pi_{ij}^{v} x_{ij}^{v} \;\; - \!\!\!\! \sum_{{\scriptsize \begin{array}{c} v \in V\\(i,j) \in \tilde{A}(v) \end{array}}} \!\!\!\!\!\!\!\!  \phi^{v} \pi_{ij}^{v} \frac{d_{ij}^{v}}{d_{0}^{v}} - \sum_{{\scriptsize \begin{array}{c} v \in V\\c \in C^v\end{array}}} \sigma^{v}_{c} s^{c}
\end{flalign}
\begin{flalign}\nonumber
  \hspace{-5cm}  s.t.:
\end{flalign}\vspace{-0.2in}
\begin{flalign}\label{BFP}
    \sum_{w \in W^{v}} p_{w}^{c} + s^{c} = g^c, \hspace{3mm} \forall v \in V, \; c \in C^v
\end{flalign}\vspace{-0.2in}
\begin{flalign}\label{BFRB}
    \sum_{j \in N_{i}^{+}(v)} \!\!\!\! x_{ji}^{v} \;\; - \sum_{j \in N_{i}^{-}(v)} \!\!\!\! x_{ij}^{v} = \left \{\begin{array}{ll}
                 1 & \textrm{if} \;\; i \in O(v)\\
                -1 & \textrm{if} \;\; i \in D(v)\\
                 0 & \textrm{otherwise}
             \end{array}\right., \hspace{3mm} \forall v \in V, \; i \in N(v)
\end{flalign}\vspace{-0.2in}
\begin{flalign}\label{BCB}
     \sum_{j \in \tilde{N}_{i}^{+}(v)} \!\!\!\! d_{ji}^{v} - \!\!\!\! \sum_{j \in \tilde{N}_{i}^{-}(v)} \!\!\!\! d_{ij}^{v} = \left \{\begin{array}{l}
                 {\displaystyle - \Delta^{p} \sum_{c \in C_{w}^{v}} p_{w}^{c}} \;\; \textrm{if} \;\; w \in W^{v}_{i}\\
                  {\displaystyle \Delta^{p} \sum_{c \in C^v}  (g^{c} - s^{c})} \;\; \textrm{otherwise}
             \end{array}\right. \hspace{1.5mm} \forall v \in V, \; i \in \tilde{N}(v)
\end{flalign}\vspace{-0.2in}
\begin{flalign}\label{LC1}
    d_{ij}^{v} \leq M_{1} \; x_{ij}^{v}, \hspace{3mm} \forall v \in V, \; (i,j) \in \tilde{A}(v)
\end{flalign}\vspace{-0.2in}
\begin{flalign}\label{LC2}
    p_{w}^{c} \leq M_{2} \!\!\!\! \sum_{j \in N^{-}_{i}(v)} \!\!\!\! x_{ji}^{v}, \hspace{3mm} \forall v \in V, \; c \in C^{v}, \; w \in W^v, \; i \in N^w
\end{flalign}\vspace{-0.2in}
\begin{flalign}\label{TVC1}
    t_{j}^{v} \geq t_{i}^{v} + \pi_{ij}^v - M_{3} (1 - x_{ij}^{v}), \hspace{3mm} \forall v \in V , \; (i,j) \in A(v): i \in O(v)
\end{flalign}\vspace{-0.2in}
\begin{flalign}\label{TVC2}
    \!\!\!\!\!\! t_{j}^{v} \geq t_{i}^{v} + \sum_{c \in C_{w}^{v}} \gamma_{w}^{p} \; p_{w}^{c} + \pi_{ij}^v - M_{4} (1 - x_{ij}^{v}), \hspace{1.5mm} \forall v \in V , \; (i,j) \in A(v), \; w \in W_{i}^v
\end{flalign}\vspace{-0.2in}
\begin{flalign}\label{TVS1}
     \!\!\!\!\!\!\!\!\!\! t_{i}^{v_{2}} \geq t_{i}^{v_{1}} + \!\! \sum_{c \in C_{w}^{v_{1}}} \!\! \gamma_{w}^{p} \; p_{w}^{c} - M_{5} (1 - y^{v_{1},v_{2}}_{w}), \hspace{1.5mm} \forall v_{1}, v_{2} \in V, \; w \in W^{v_{1},v_{2}}, \; i \in N^w
\end{flalign}\vspace{-0.2in}
\begin{flalign}\label{TVS2}
    \!\!\!\!\!\!\!\!\!\! \sum_{j \in N_{i}^{+}(v_{1})} \!\!\!\!\!\! x_{ji}^{v_1} + \!\!\!\!\!\!\!\! \sum_{j \in N_{i}^{+}(v_{2})} \!\!\!\!\!\! x_{ji}^{v_2} \leq y_{w}^{v_{1},v_{2}} + y_{w}^{v_{2}, v_{1}} + 1, \hspace{1.5mm} \forall v_{1}, v_{2} \in V, \; w \in W^{v_{1},v_{2}}, \; i \in N^w
\end{flalign}\vspace{-0.2in}
\begin{flalign} \label{LCS}
    \sum_{j \in \tilde{N}^{+}_{i}(v)} \!\!\!\! d_{ji}^{v} + \Delta^{p} \sum_{c \in C_{w}^{v}} p_{w}^{c} \leq \Delta^{v}_{w}, \hspace{1.5mm} \forall v \in V, \; w \in W^v, \; i \in N^w
\end{flalign}\vspace{-0.2in}
\begin{flalign}\label{VH1}
   t_{i}^{v} \geq l_{w} \!\!\!\! \sum_{j \in N^{-}_{i}(v)} \!\!\!\! x_{ji}^{v}, \hspace{3mm} \forall v \in V, \; w \in W^{v}, \; i \in N^w
\end{flalign}\vspace{-0.2in}
\begin{flalign}\label{VH2}
   t_{i}^{v} + \sum_{c \in C_{w}^{v}} \gamma_{w}^{p} \; p_{w}^{c} \leq u_{w}, \hspace{3mm} \forall v \in V, \; w \in W^{v}, \; i \in N^w
\end{flalign}\vspace{-0.2in}
\begin{flalign}\label{VDD}
   t_{i}^v \leq \xi^c, \hspace{3mm} \forall v \in V, \; c \in C^v, \; i \in D(v)
\end{flalign}


The objective function \eqref{FO} maximizes the benefit of the tramp shipping company given the total income $\psi_{w}^{c}$ from the pallets transported $p^w_c$ (first term) and the costs incurred given by the fixed fares $\theta_{w}^{v}$ associated with the berth time windows where vessels operate $x_{ji}^{v}$ (second term), the vessel renting cost $\rho^{v}$ (third term), the fuel consumption $\phi_{v}$  throughout the vessel route (fourth and fifth terms), and the compensation expenses $\sigma_{w}^{c}$ incurred due to pallets not carried (six term). The fourth term captures the minimum fuel consumption cost according to the light draft of vessels (without load) on each sailing leg ($x_{ij}^{v} = 1$); whereas the fifth term considers the extra fuel cost when carrying some load according to the ratio between the vessel draft variation and the light draft of the vessel ($\frac{d_{ij}^{v}}{d_{0}^{v}}$). Constraints \eqref{BFP} perform the balance of loaded pallets in each contract. This constraint allows partially fulfilling a contract, i.e., some pallets may not be loaded in any vessel $s^c$. Constraints \eqref{BFRB} represent the ship flow balance where the source node is associated with the origin of the vessel $O(v)$ and the sink node corresponds to the destination of the vessel $D(v)$. Thus, the rest of nodes are intermediate nodes $N(v)$ where flow conservation is guaranteed. Constraints \eqref{BCB} perform the balance of the draft variation of vessels  throughout a subnetwork where only links connecting pairs of expanded berth nodes, as well as those links connecting the expanded berth nodes with the destination nodes of the contracts are considered. When balancing the flow in a node, two situations may arise. If it is a berth time window ($w \in W_{i}^v$), a draft increment is done. This increment is proportional to the number of loaded pallets in the vessel at the time window, and is calculated using parameter $\Delta^{p}$. Otherwise, it is a destination of some contracts and a draft decrement is performed. This decrement is also proportional to the number of transported pallets from the contracts having this destination. Constraints \eqref{LC1} - \eqref{LC2} link the ship routing variables ($x_{ij}^{v}$) to draft variation flows ($d_{ij}^{v}$), and to the amounts of loaded pallets in each berth time windows ($p_{w}^{c}$), respectively. Large constants $M_{1}$ and $M_{2}$ limit the maximum values that decision variables $d_{ij}^{v}$ and $p_{w}^{c}$ can take. The former takes into account the maximum allowable draft of the berth associated with the time windows; whereas the latter considers the total amount of pallets to be loaded according to the established contract. Constraints \eqref{TVC1}-\eqref{TVC2} establish the schedule of two nodes $(i,j) \in A(v)$ when visited consecutively by the same vessel (i.e., $x_{ij}^{v} = 1$). In both sets of constraints, the sailing time between these pair of nodes ($\pi_{ij}^{v}$) is considered. In \eqref{TVC2}, the time spent for loading pallets in the vessel in node $i$ is also considered if this node is related to a berth time windows $w \in W_{i}^v$. Large constants $M_{3}$-$M_{4}$ disables these constraints when vessel $v$ does not sail between nodes $i, j$ (i.e., $x_{ij}^{v} = 0$). Constraints \eqref{TVS1} - \eqref{TVS2} coordinate the scheduling of two vessels $v_1, v_{2} \in V$ when operating in the same time window $w \in W^{v_{1},v_{2}}$. If vessel $v_{1}$ docks first ($y^{v_{1},v_{2}}_{w} = 1$) then (\ref{TVS1}) ensures that vessel $v_2$ docks after all pallets of $v_1$ have been loaded. Otherwise, $y^{v_{1},v_{2}}_{w} = 0$, and large constant $M_{5}$ disables this constraint. The value of $y^{v_{1},v_{2}}_{w}$ is set by \eqref{TVS2}. Constraints \eqref{LCS} ensure that the vessels draft variation does not exceed the allowable maximum variation for each berth time window in which the vessel operates. The vessel's draft variation is measured according to the current transit draft variation through the incoming link plus the draft increment due to pallet loading in the berth time window. Constraints \eqref{VH1} - \eqref{VH2} guarantee that vessels operate within the available time windows of berths. Constraints (\ref{VH1}) verify that vessels do not operate before than the lower bound of the time windows, whereas (\ref{VH2}) do not allow vessels to stay longer than the upper bound of this time window. The latter also considers loading pallet time. Finally, constraints \eqref{VDD} make sure that vessels arrive at the destination ports of the cargoes not later than the due date established in their contracts.

\subsection*{Time windows reduction rules}

Information of berth time windows can be used to reduce problem size, thus enhancing solution efficiency. The narrower the time window the larger the number of binary variables and sequencing constraints that can be removed from the problem formulation. The time window-based elimination rules that consider the time windows as hard constraints are inspired on \cite{DC07}. In this way, a vessel is prevented from using a time window if there is not enough time in for arrive¿ing and loading pallets. In this case, binary variable $x_{ij}^{v}$ and constraints \eqref{LC1} and \eqref{TVC2} can all be dropped from the problem formulation. Also, a time window is prevented from being used if the service time remaining at the berth when a vessel arrives from its origin node is not enough to load a minimum amount of pallets. Therefore, binary variable $x_{ij}^{v}$ and constraints \eqref{TVC1} can be eliminated from the model.

\section{Solution Approach}\label{sSSA}

\begin{figure}[h]
\begin{center}
\includegraphics[scale=1]{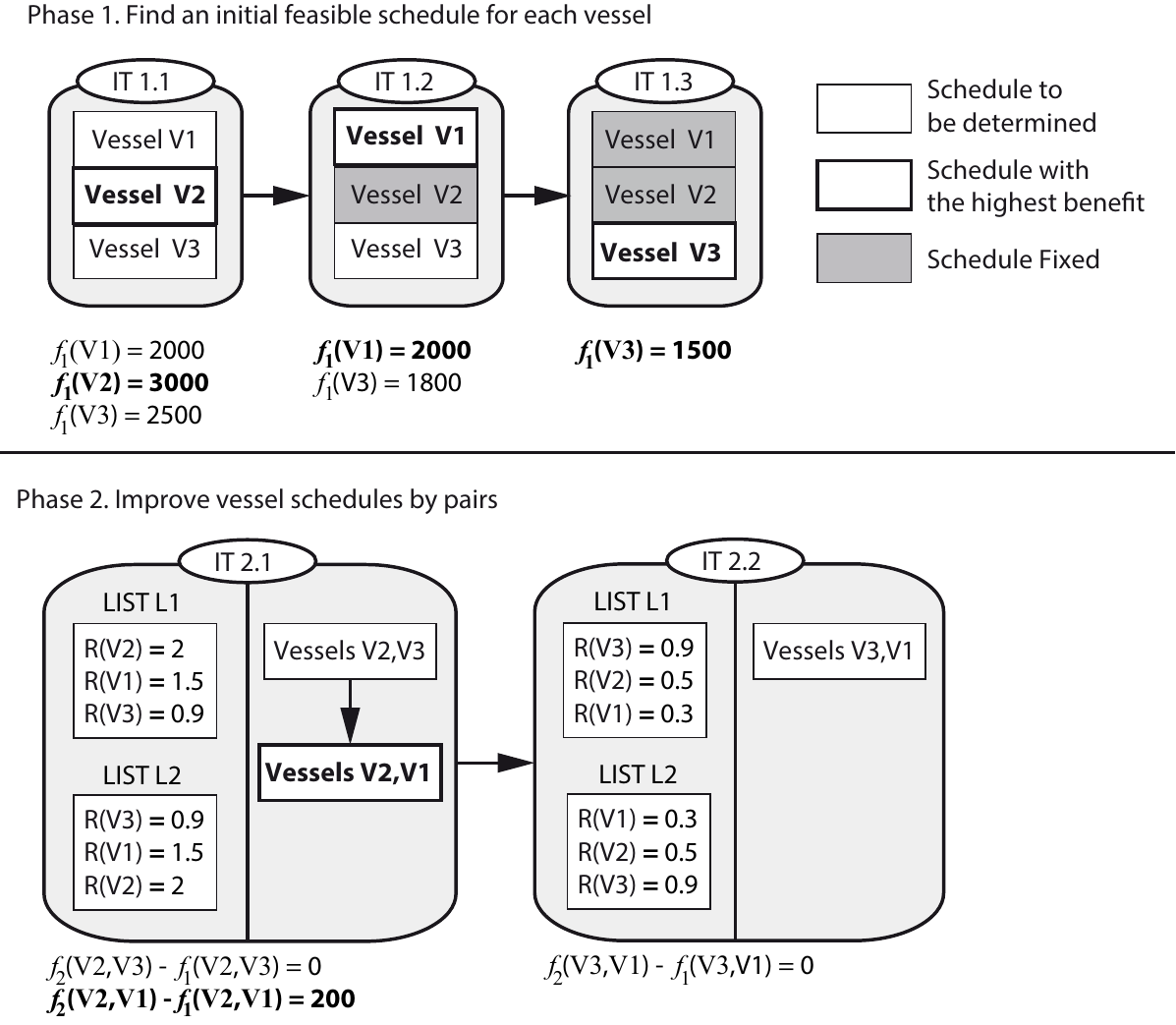}
\caption{An illustrative example for the 2-phase heuristic consisting of three vessels.}\label{fSSA}
\end{center}
\end{figure}

The MILP model in this work becomes harder to solve as the number of vessels increases. All the decision variables and constraints of the model are indexed over the set of vessels. Moreover, constraints \eqref{TVS1}-\eqref{TVS2} link decisions between pairs of vessels and, thus, the schedules of the vessels cannot be determined separately. Therefore, a heuristic decomposition method has been developed for solving efficiently instances with several vessels. As shown in Fig. \ref{fSSA}, the method is divided into two phases. Phase 1 determines quickly an initial schedule for each vessel, whereas phase 2 tries to improve/verify that these schedules are (near-)optimal while not spending too much time.

The algorithm for the first phase is described in table \ref{tbAlgP1} and works as follows. Initially, constraints \eqref{TVS1}-\eqref{TVS2} are dropped from the model, and tentative schedules for each vessel are computed separately. Then, the schedule of the vessel with the highest benefit is fixed and constraints \eqref{TVS1}-\eqref{TVS2} related to this vessel are released. Next, the rest of schedules are again computed and the following vessel schedule with the highest benefit is fixed. This procedure is repeated as long as two or more vessel schedules have not been fixed. Finally, the schedule of the vessel with the lowest benefit is determined. The time spent in this phase can be dramatically decreased if the procedure is implemented in parallel, i.e., in each iteration (IT), the computation of each vessel schedule can be done in a different CPU, so that the whole iteration time would be the highest time spent in solving one single vessel schedule.

\begin{table}[!]
\begin{center}
\begin{tabular}{|l|}\hline
1. \textbf{Set} $\tilde{V} \leftarrow V$;\\[0.05in]
2. \textbf{Drop} constraints \eqref{TVS1} - \eqref{TVS2} from the model;\\[0.05in]
3. \textbf{Compute} schedules $\forall v \in \tilde{V}$;\\[0.05in]
4. \textbf{If} $|\tilde{V}| = 1$ \textbf{then} \textbf{return} Initial Schedules;\\[0.05in]
5. \textbf{Fix} schedule of $\tilde{v} \in \tilde{V}: f(\tilde{v}) = \max_{v \in \tilde{V}} f(v)$;\\[0.05in]
6. \textbf{Update} $\tilde{V} \leftarrow \tilde{V} \backslash \{\tilde{v}\}$;\\[0.05in]
7. \textbf{Release} constraints \eqref{TVS1} - \eqref{TVS2} in which $\tilde{v}$ is involved \textbf{and} \textbf{goto} 3;\\\hline
\end{tabular}
\end{center}
\caption{Algorithm for the Phase 1 of the heuristic.}\label{tbAlgP1}
\end{table}

An illustrative application of this phase is shown on the top of Fig. \ref{fSSA}. In that application, the initial schedules of three vessels labeled as V1, V2 and V3 are determined. In the first iteration (IT 1.1), the schedules of each vessel are computed in parallel, having benefits ($f_{1}(Vi)$) of 2000, 3000 and 2500, respectively. As $f_{1}(V2)$ is the highest one, then the schedule of V2 is fixed in iteration 2 (IT1.2), so the schedules of V1 and V3 are re-optimized, having new benefits of 2000 and 1800, respectively. $f_{1}(V1)$ is now the highest benefit of the remaining vessels, so the schedule of V1 is fixed in iteration 3 (IT1.3), where only the new schedule of V3 is determined.

Phase 2 tries to improve the initial vessel schedules determined in phase 1. In this phase, the schedules of pair of vessels are re-optimized in the hope that the two new schedules may lead to a better global solution, although the solution for the schedule of one vessel in the pair may be worse. The total number of pair evaluations increases dramatically with the number of vessels, therefore a mechanism for evaluating a reduced number of these has been developed. By comparing the solutions provided by the first phase and the resolution of the full model, a suitable selection of pair of vessels can be determined by using a ratio measure. This ratio evaluates the relationship between the remaining capacity of the vessel ($\Delta Q$) and the average remaining bandwidth of the berth time windows ($\Delta\overline{B}$). $\Delta Q$ is computed by reducing the vessel capacity ($Q$) according to the number of pallets loaded in the vessel in each time window; whereas $\Delta\overline{B}$ is obtained in three steps. First, the bandwidth of each time window $w \in W$ ($B_w$) is calculated as the difference between its upper and lower bounds ($u_{w}$ and $l_{w}$, respectively). Then, each $B_w$ in which the vessel has operated is reduced according to the time spent in loading pallets in the vessel in that time window. Finally, all $B_w$ are averaged yielding to $\Delta\overline{B}$. Experimentation practice on small-sized instances, where optimal schedules are achieved by solving directly the model \eqref{FO}-\eqref{VDD}, shows that re-optimizing schedules of vessel pairs where one vessel has a high ratio and another a low one leads also to the optimal schedules.

The procedure of this phase is shown in table \ref{tbAlgP2} and works as follows. Initially, the ratio measure of each vessel ($R(v)$) is first computed, and then two lists (L1 and L2) are built. In list L1, vessels are sorted in descending ratio, whereas in L2 vessels are ordered in an increasing fashion. Next, the algorithm goes into an iteration process working as follows. A vessel is picked up from each list conforming a pair that has not been previously evaluated ($\tau(L1(p_{1}), L2(p_{2})) = 0$), and that the ratio of the vessel from L1 is greater than the one from L2 ($R(L1(p_{1})) > R(L2(p_{2}))$). The schedules of each pair of vessels meeting these requirements are re-optimized as long as no improvements are found, and all valid pairs are not evaluated. Once the benefit of a pair ($f(L1(p_{1}),L2(p_{2}))$) is improved, the iteration process is terminated, and another iteration process is started by updating the ratios of the vessels associated with that pair as well as lists L1 and L2. The procedure is terminated when all pairs of valid vessels have been evaluated.

\begin{table}[!]
\begin{center}
\begin{tabular}{|l|}\hline
1. \textbf{Set} $\tilde{f}(v) \leftarrow f(v)$  $\forall v \in V$, \textbf{and} $\tau(v_{i}, v_{j}) \leftarrow 0$  $\forall v_{i}, v_{j} \in V$;\\[0.05in]
2. \textbf{Compute} $R(v)$  $\forall v \in V$;\\[0.05in]
3. \textbf{Set} L1 $\leftarrow v \in V$ \textbf{sorted by} $R(v)$ in \textbf{decreasing} fashion;\\[0.05in]
\hspace{9mm} L2 $\leftarrow v \in V$ \textbf{sorted by} $R(v)$ in \textbf{ascending} fashion;\\[0.05in]
4. \textbf{Set} $p_{1} \leftarrow 1$, \; $p_{2} \leftarrow 1$;\\[0.05in]
5. \textbf{If} $\tau(L1(p_{1}), L2(p_{2})) = 1$ \textbf{or} $R(L1(p_{1})) < R(L2(p_{2}))$ \textbf{then} \textbf{jump to} 10;\\[0.05in]
6. \textbf{Compute} schedules for pair $(L1(p_{1}), L2(p_{2}))$;\\[0.05in]
7. \textbf{If} $f(L1(p_{1}),L2(p_{2})) > \tilde{f}(L1(p_{1}),L2(p_{2}))$ \textbf{then} \textbf{jump to} 10;\\[0.05in]
8. \textbf{Set} $\tilde{f}(L1(p_{1})) \leftarrow f(L1(p_{1}))$, \; $\tilde{f}(L2(p_{2})) \leftarrow f(L2(p_{2}))$ \textbf{and} $\tau(L1(p_{1}), L2(p_{2})) \leftarrow 0$;\\[0.05in]
9. \textbf{Update} $R(L1(p_{1}))$, $R(L2(p_{2}))$ \textbf{and} \textbf{goto} 3;\\[0.05in]
10. \textbf{If} $p_{2} < |L2|$ \textbf{then} $p_{2} \leftarrow p_{2} + 1$, \textbf{and} \textbf{goto} 5;\\[0.05in]
11. \textbf{If} $p_{1} < |L1|$ \textbf{then} $p_{1} \leftarrow p_{1} + 1$, \textbf{and} \textbf{goto} 5;\\[0.05in]
12. \textbf{Return} Improved schedules;\\\hline
\end{tabular}
\end{center}
\caption{Algorithm for the Phase 2 of the heuristic.}\label{tbAlgP2}
\end{table}

An illustrative application of phase 2 is shown on the bottom of Fig. \ref{fSSA}. On this figure, improved schedules for the initial schedules obtained in the application example of phase 1 are sought after. Initially, the ratio measures of vessels $V1$, $V2$ and $V3$ ($R(V1)$, $R(V2)$ and $R(V3)$, respectively) are computed. Then, lists $L1$ and $L2$ are built and valid pairs of vessels are evaluated in the first iteration process (IT 2.1.). Vessels V2 and V3 are in the top of each list and $R(V2) > R(V3)$, so the schedules for this pair are re-optimized. As the benefit of the pair is not improved (i.e., $f_2(V2,V3) = f_1(V2,V3)$), the following pair is evaluated. This corresponds to $(V2, V1)$ since $V1$ is next to $V3$ in $L2$ and $R(V2) > R(V1)$. Having re-optimized pair $(V2,V1)$, an improved benefit of 200 is obtained. Therefore, ratios of $R(V2)$ and $R(V1)$ are updated, as well as lists L1 and L2, and the second iteration process(IT 2.2) is started. Now, the pair $(V3,V1)$ is evaluated as vessels $V3$ and $V1$ are in the top of lists L1 and L2, and $R(V3) > R(V1)$. The re-optimization of this pair does not improve the benefit (i.e., $f_2(V3,V1) = f_1(V3,V1)$), and as no further valid pairs are found (i.e., pairs $(V3, V2)$ and $(V2,V1)$ have already been evaluated in IT 2.1.), phase is terminated.

\section{Results}\label{sRP}

To evaluate the model as well as the performance of the heuristic, several experiments based on a real case have been conducted. The model and the heuristic have been coded in AMPL, and the Branch and Bound of CPLEX v12.5.0.0 is used for solving the model, as well as the reduced models constructed during the heuristic execution, under a workstation R5500 with processor Intel(R) Xeon(R) CPU E5645 2.40 GHz and 48 GBytes of RAM. 30 minutes of time limit for the full model and for the heuristic has been set since the availability of time windows may change afterwards, and thus the provided solution would be unrealistic.

The main components of each instance includes the number of vessels that can be used, the number of contracts willing to be carried, the number of berths where vessels can operate, the number of available time windows per berth, and the total number of pallets that can be loaded in the vessels. Observe that, the instance identifier contains all this information except for the amount of pallets provided in Table \ref{tbSmInst}. For example, the group of instances containing the string $S4B5W2C18$ are made up of 4 vessels, 18 contracts, 5 berths and 2 time windows per berth. Detailed information on the planning process currently being used is not shown due to confidentiality issues.

\begin{table}[h!]
  \centering
    \begin{tabular}{|c|c|c|c|c|}
    \hline
    \multirow{2}[0]{*}{\textbf{Instance}} & \multicolumn{4}{c|}{\textbf{\# Pallets}} \\
    \cline{2-5}
          & \textbf{A} & \textbf{B} & \textbf{C} & \textbf{D} \\
    \hline
    \textbf{S4B5W2C18} & 22189 & 21703 & 23758 & 22741 \\
    \hline
    \textbf{S8B7W3C36} & 45132 & 43381 & 45472 & 45462 \\
    \hline
    \textbf{S12B10W3C54} & 67141 & 68910 & 66995 & 68027 \\
    \hline
    \textbf{S16B24W3C72} & 89759 & 90291 & 92184 & 91359 \\
    \hline
    \textbf{S20B30W3C90} & 111755 & 111292 & 112816 & 109642 \\
    \hline
\end{tabular}
  \caption{Characterization of the performed experiments}\label{tbSmInst}
\end{table}

Table \ref{tbPerfRes} shows the performance results for the tested experiments. From left to right of the table, it is given the best objective functions ($f$), the relative gaps in percentage (GAP($\%$)), and the computational times in seconds (T$_{CPU}$) for the direct resolution of the model (FM), and each phase of the heuristic. H1 stands for the first phase of the heuristic, which is implemented in parallel, whereas H2 denotes the second phase. Finally, HT represents the whole time spent by the heuristic.

\begin{table}[h!]
  \centering
  \begin{tabular}{|l||r|r|r||r|r||r|r|r|r|}\hline
  \multicolumn{1}{|c||}{Instance} & \multicolumn{3}{c||}{$f$(USD x $10^3$)} & \multicolumn{2}{c||}{GAP($\%$)} & \multicolumn{4}{c|}{T$_{CPU}$(secs)}\\ \cline{2-10}
      & \multicolumn{1}{c|}{FM} & \multicolumn{1}{c|}{H1} & \multicolumn{1}{c||}{H2} & \multicolumn{1}{c|}{H1} & \multicolumn{1}{c||}{H2} & \multicolumn{1}{c|}{FM} & \multicolumn{1}{c|}{H1} & \multicolumn{1}{c|}{H2} & \multicolumn{1}{c|}{HT}\\ \hline \hline
  S4B5W2C18A & 1414 & 1397 & 1414 & 1.2 $\%$ & 0 $\%$ & 10 & 1 & 5 & 6\\ \hline
  S4B5W2C18B & 1630 & 1630 & 1630 & 0 $\%$ & 0 $\%$ & 44 & 2 & 13 & 15\\ \hline
  S4B5W2C18C & 1404 & 1271 & 1404 & 9.5 $\%$ & 0 $\%$ & 21 & 1 & 8 & 9\\ \hline
  S4B5W2C18D & 1611 & 1599 & 1611 & 0.8 $\%$ & 0 $\%$ & 3 & 1 & 3 & 4\\ \hline \hline

  S8B7W3C36A & 2081 & 2078 & 2081 & 0.2 $\%$ & 0 $\%$ & 1800 & 3 & 44 & 47\\ \hline
  S8B7W3C36B & 1729 & 1640 & 1729 & 5 $\%$ & 0 $\%$ & 1800 & 5 & 186 & 191\\ \hline
  S8B7W3C36C & 1704 & 1704 & 1704 & 0 $\%$ & 0 $\%$ & 1800 & 3 & 65 & 68\\ \hline
  S8B7W3C36D & 2739 & 2539 & 2779 & 7 $\%$ & - 1.5 $\%$ & 1800 & 6 & 222 & 228\\ \hline \hline

  S12B10W3C54A & 3052 & 3052 & 3052 & 0 $\%$ & 0 $\%$ & 1800 & 200 & 1600 & 1800\\ \hline
  S12B10W3C54B & 2725 & 2725 & 2725 & 0 $\%$ & 0 $\%$ & 1800 & 89 & 1711 & 1800\\ \hline
  S12B10W3C54C & 2700 & 2700 & 2703 & 0 $\%$ & - 0.1 $\%$ & 1800 & 64 & 1736 & 1800\\ \hline
  S12B10W3C54D & 2634 & 2890 & 2894 & -9.8 $\%$  & -9.9 $\%$ & 1800 & 99 & 1701 & 1800\\ \hline

  S16B24W3C72A & 3829 & 4122 & 4129 & -7.6 $\%$ & -7.8 $\%$ & 1800 & 303 & 1697 & 1800\\ \hline
  S16B24W3C72B & 3745 & 3788 & 3791 & -1.1 $\%$ & -1.2 $\%$ & 1800 & 255 & 1745 & 1800\\ \hline
  S16B24W3C72C & 3748 & 3937 & 3938 & -5 $\%$ & -5 $\%$ & 1800 & 256 & 1744 & 1800\\ \hline
  S16B24W3C72D & 3848 & 3911 & 3914 & -1.6 $\%$ & -1.7 $\%$ & 1800 & 265 & 1735 & 1800\\ \hline

  S20B30W3C90A & 4467 & 4467 & 4562 & 0 $\%$ & -2 $\%$ & 1800 & 554 & 1246 & 1800\\ \hline
  S20B30W3C90B & 4855 & 5027 & 5034 & -3.5 $\%$ & -3.7 $\%$ & 1800 & 299 & 1501 & 1800\\ \hline
  S20B30W3C90C & 4519 & 4529 & 4529 & -0.2 $\%$ & -0.2 $\%$ & 1800 & 591 & 1209 & 1800\\ \hline
  S20B30W3C90D & 4300 & 4369 & 4376 & -1.6 $\%$ & -1.8 $\%$ & 1800 & 477 & 1323 & 1800\\ \hline
\end{tabular}
  \caption{Summary results for the performed experiments.}\label{tbPerfRes}
\end{table}

Results show that the heuristic outperforms the resolution of the full model. For small-sized instances (4 vessels), the heuristic finds always the optimum having improved the initial solution. For medium-sized instances (8 vessels), the heuristic finds at least the same feasible solution (this solution is not guaranteed to be optimal since the resolution of the full model have been earlier stopped, having reached the time limit of 30 minutes). Moreover, the heuristic only spends a few minutes. Finally, for larger instances (12-20 vessels), the heuristic provides better solutions, and quite frequently in early phase 1. In some cases, phase 2 improves the initial solution but not substantially. This is because solving times reach the time limit and, thus, phase 2 is terminated earlier.

Next figure \ref{fEx} shows the workings of the heuristic for instance S4B5W2C18C, the instance with the highest improvement obtained in phase 2.
It can be observed that in phase 1, there is no schedule determined for vessel 1 and that the rest of vessels only dock once and in a different time window. However, by coordinating arrivals at the same time window and by docking twice, phase 2 obtains better schedules (i.e., benefits are higher) and allows vessel 1 to operate.

\begin{figure}[h!]
\begin{center}
\includegraphics[scale=1]{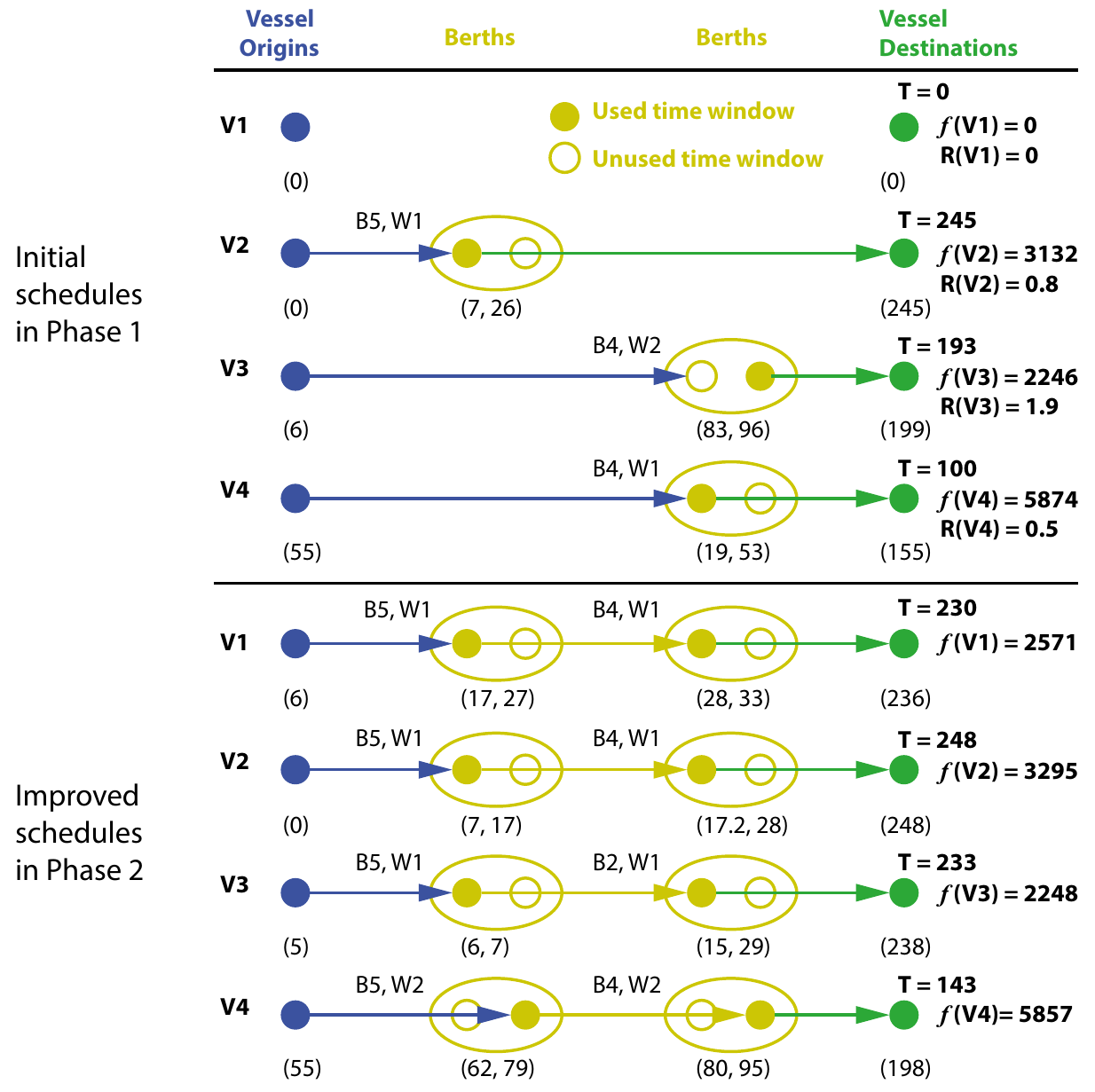}
\caption{Schedules determined in phases 1 and 2 of the heuristic for instance S4B5W2C18C.}\label{fEx}
\end{center}
\end{figure}

Finally, table \ref{tbSolRes} provides some details of the solution results. From left to right of the table, it is provided the average and maximum values for the number of docks (\# Docks), the average used capacity of the vessels (Av. Used Cap.), and the percentage of cargo satisfaction (Cargo Satisfied (\%)) for the direct resolution of the model and each phase of the heuristic.

\begin{table}[h!]
  \centering
  \begin{tabular}{|l||r|r|r||r|r|r||r|r|r|}\hline
  \multicolumn{1}{|c||}{Instance} & \multicolumn{3}{c||}{\# Docks (Av-Max)} & \multicolumn{3}{c||}{Av. Used Cap.} & \multicolumn{3}{c|}{Cargo Satisfied}\\ \cline{2-10}
      & \multicolumn{1}{c|}{FM} & \multicolumn{1}{c|}{H1} & \multicolumn{1}{c||}{H2} & \multicolumn{1}{c|}{FM} & \multicolumn{1}{c|}{H1} & \multicolumn{1}{c||}{H2}  & \multicolumn{1}{c|}{FM} & \multicolumn{1}{c|}{H1} & \multicolumn{1}{c|}{H2} \\ \hline \hline
  S4B5W2C18A & 1.6 - 3 & 1.6 - 2 & 1.6 - 3 & 41\% & 40\% & 41\% & 57\% & 56\% & 57\%\\ \hline
  S4B5W2C18B & 1.6 - 2 & 1.6 - 2 & 1.6 - 2 & 45\% & 45\% & 45\% & 57\% & 57\% & 57\%\\ \hline
  S4B5W2C18C & 2 - 2 & 1 - 1 & 2 - 2 & 51\% & 46\% & 51\% & 50\% & 47\% & 50\%\\ \hline
  S4B5W2C18D & 2 - 2 & 1.6 - 2 &  2 - 2 & 45\% & 44\% & 45\% & 58\% & 57\% & 58\%\\ \hline \hline

  S8B7W3C36A & 1.5 - 2 & 1.5 - 2 & 1.5 - 2 & 48\% & 48\% & 48\% & 58\% & 58\% & 58\%\\ \hline
  S8B7W3C36B & 1 - 1 & 1 - 1 & 1 - 1 & 27\% & 25\% & 27\% & 60\% & 57\% & 60\% \\ \hline
  S8B7W3C36C & 1.2 - 2 & 1.2 - 2 & 1.2 - 2 & 31\% & 31\% & 31\% & 53\% & 53\% & 53\%\\ \hline
  S8B7W3C36D & 1.8 - 2 & 1.3 - 2 & 2 - 3 & 41\% & 38\% & 42\% & 35\% & 32\% & 35\%\\ \hline \hline

  S12B10W3C54A & 1.3 - 2 & 1.3 - 2 & 1.3 - 2 & 46\% & 46\% & 46\% & 46\% & 46\% & 46\%\\ \hline
  S12B10W3C54B & 1 - 1 & 1 - 1 & 1 - 1 & 41\% & 41\% & 41\% & 54\% & 54\% & 54\%\\ \hline
  S12B10W3C54C & 1 - 1 & 1 - 1 & 1 - 1 & 31\% & 31\% & 31\% & 57\% & 57\% & 57\%\\ \hline
  S12B10W3C54D & 1.3 - 3 & 1.3 - 2 & 1.3 - 2 & 39\% & 45\% & 45\% & 33\% & 40\% & 40\%\\ \hline

  S16B24W3C72A & 1.3 - 2 & 1.3 - 2 & 1.3 - 2 & 39\% & 70\% & 70\% & 26\% & 64\% & 64\%\\ \hline
  S16B24W3C72B & 1.8 - 3 & 1.4 - 3 & 1.4 - 3 & 46\% & 48\% & 48\% & 46\% & 48\% & 48\%\\ \hline
  S16B24W3C72C & 1 - 2 & 1.2 - 2 &  1.3 - 2 & 35\% & 43\% & 43\% & 36\% & 43\% & 44\%\\ \hline
  S16B24W3C72D & 1 - 2 & 1.3 - 2 & 1.2 - 2  & 48\% & 51\% & 51\% & 46\% & 49\% & 40\%\\ \hline

  S20B30W3C90A & 1.4 - 3 & 1.3 - 3 & 1.5 - 3 & 40\% & 43\% & 43\% & 45\% & 48\% & 48\%\\ \hline
  S20B30W3C90B & 1.3 - 3 & 1.3 - 3 & 1.3 - 3 & 46\% & 48\% & 48\% & 47\% & 48\% & 48\%\\ \hline
  S20B30W3C90C & 1.3 - 2 & 1.2 - 2 & 1.2 - 2  & 52\% & 53\% & 53\% & 56\% & 57\% & 57\%\\ \hline
  S20B30W3C90D & 1.4 - 2 & 1.4 - 2 & 1.2 - 2 & 25\% & 25\% & 27\% & 37\% & 37\% & 38\%\\ \hline
\end{tabular}
  \caption{Some features of the solution results for the performed experiments.}\label{tbSolRes}
\end{table}

Results show that vessel capacities are underused, around 50\% on average, and with minimum and maximum values of 26 \% and 64\%, respectively. Another important aspect to highlight is that vessels do not dock more than 3 times, an issue that also happens in real practice. However, the average number of docks is around 1.5, meaning that in the schedule solution one half of the vessels docks once and the other half twice. Regarding to the amounts of carried pallets, they are low but the company has benefit as shown in the previous table. In reality, the company would have important losses during the fruit season if all cargoes were carried, so lower benefits are obtained since the time windows at berths are narrower and operational costs are difficult to absorb.

\section{Conclusions and Further Research}\label{sCFR}

This work introduces an optimisation model that deals with the tramp ship scheduling problem considering berth allocation issues. The model is formulated as a mixed-integer linear problem and carries out the assignment of berth time windows to vessels where cargoes are loaded, the itineraries of the vessels with the possibility of docking several times at the same or different berth, and the coordination of vessels operating at the same berth time window. Moreover, the model takes into account berth limitations regarding to water depth or draft capacity. These limitations have not been previously addressed on the scant literature integrating the berth allocation to the tramp ship scheduling problem.

From a business practice perspective, the proposed approach provides a decision-support tool for the tramp shipping company. Current planning activities are done manually by operations planners, who find themselves with several difficulties in the search of a feasible plan. These difficulties arise when trying to match temporal availability, draft capacities and scheduling issues at the same time. These aspects need a precise coordination for a schedule to be feasible, and this is the cause for the success of this modeling approach when dealing with the tramp ship scheduling with berth allocation issues. Furthermore, this optimization approach allows operation planners to perform scenario analysis by varying different problem aspects such as cargo size, loading times, time window fare or draft, among others.

For further research, several extensions can be considered for the tramp ship scheduling problem. In this regard, we propose to incorporate as a decision variable the assignment of cargo to vessels, which is currently not considered in the mathematical model motivated by the fact that the shipping company assumes that this is a decision taken by the commercial department of the company. However, integrating this decision variable could lead to better solutions and reduce overall costs. Another potential extension is related to speed optimization, which currently is assumed as a fixed value. This can be done by incorporating an emission estimation model in order to account for environmental concerns. Another research avenue to extend the proposed model is to deal with risk and uncertainty as modelling elements and provide an optimization framework to deal with disruption events in the day-to-day operations. Finally, because of the interaction between berth allocation decisions of the stevedores and the shipping company, we propose to formulate the problem as a bi-level optimization model in which the port terminal managers can be the leader that determines the berth assignment and schedules, and the shipping company the follower that receives this information as berth availability time windows. Other integration approaches can be also considered such as a pre-processing or a feedback loop can be also explored.

\section*{Acknowledgements}

This work was supported by the Mexican National Council for Science and Technology (CONACYT), through research grant SEP-CONACYTCB-2014-01-240814 (third author); and by the Vice Presidency of Research and Graduate Studies of Pontificia Universidad Cat\'olica de Valpara\'iso, through research project 037.499/2015 (second author). Additionally, we acknowledge the determination and effort performed by the Mexican Logistics and Supply Chain Association (AML) and the Mexican Institute of Transportation (IMT) for providing us an internationally recognized collaboration platform. Last, but not least, special thanks to the undergraduate students, David Hirmas and Marco Sanhueza from the Pontificia Universidad Cat\'olica de Valpara\'iso, for their help in developing computational experiments, and an anonymous planner of the shipping company who allowed us to fully understand the complex planning process of the company.

\bibliographystyle{plain}

\begin{thebibliography}{99}

\bibitem[Bierwirth and Meisel(2015)]{BM15} Bierwirth C, Meisel F. (2015). A follow-up survey of berth allocation and quay crane scheduling problems in container terminals. Eur J Oper Res, 244, 675-689.



\bibitem[Christiansen et al.(2013)]{CFNR13} Christiansen M, Fagerholt K, Nygreen B, Ronen D. (2013). Ship Routing and scheduling in the new milenium. Eur J Oper Res, 228, 467-483.

\bibitem[Dondo and Cerd\'a(2007)]{DC07} Dondo R, Cerd\'a J. (2007). A cluster-based optimization approach for the multi-depot heterogeneous fleet vehicle routing problem with time windows. Eur J Oper Res, 176, 1478-1507.




\bibitem[Fagerholt and Ronen(2013)]{FR13}Fagerholt K, Ronen D. (2013). Bulk ship routing and scheduling: Solving practical problems may provide better results. Maritime Policy \& Management, 40(1): 48-64.


\bibitem[Hennig et al.(2012)]{HFKN12} Hennig F, Nygreen B, Christiansen M, Fagerholt K, Furman KC, Song J, Kocis GR, Warrick PH. (2012). Maritime crude oil transportation - a split pickup and split delivery problem. Eur J Oper Res 218, 764-774.


\bibitem[Jetlund and Karimi (2004)]{JK04} Jetlund AS, Karimi IA. (2004). Improving the logistics of multi-compartment chemical tankers. Comput Chem Eng, 28, 1267-1283.

\bibitem[Korsvik and Fagerholt(2010)]{KF10} Korsvik JE, Fagerholt K. (2010). A tabu search heuristic for ship routing and scheduling with flexible cargo quantities. J Heuristics 16, 117-137.

\bibitem[Korsvik et al.(2011)]{KFL11} Korsvik JE, Fagerholt K, Laporte G. (2011). A large neighbourhood search heuristic for ship routing and scheduling with split loads. Comput Oper Res 38, 474-483


\bibitem[Li and Pang(2011)]{LP11} Li C-L, Pang K-W.(2011). An integrated model for ship routing and berth allocation. Int J Shipp \& Transp Logist, 3(3), 245-260.

\bibitem[Pang and Liu(2014)]{PL14} Pang K-W, Liu J. (2014). An integrated model for ship routing with transshipment and berth allocation, IIE Trans, 46(12), 1357-1370.



\bibitem[UNCTAD(2014)]{UNCTAD04} UNCTAD secretariat. (2014). Maritime Transport Review. In: Rogers, J., (eds), United Nation Publications. Ed. UNCTAD, Geneva.

\end{thebibliography}

\end{document}